\documentclass[conference]{IEEEtran}
\usepackage{times}

\usepackage[numbers]{natbib}
\usepackage{multicol}
\usepackage[bookmarks=true]{hyperref}
\usepackage[caption=false,font=normalsize,labelfont=sf,textfont=sf]{subfig}
\usepackage{textcomp}
\usepackage{stfloats}
\usepackage{url}
\usepackage{verbatim}
\usepackage{graphicx}
\begin{document}

\title{Towards the Neuromorphic Computing for Offroad Robot Environment Perception and Navigation}

\author{Author Names Omitted for Anonymous Review. Paper-ID [105]}



%
\author{\authorblockN{Zhenhua Yu\authorrefmark{1}$^{1}$$^{2}$, Peter R.N. Childs$^{2}$, Thrishantha Nanayakkara$^{2}$} 
\vspace{-0.4cm}
\authorblockA{\\$^{1}$Department of Mechanical Engineering, Imperial College London, UK; Email: z.yu18@imperial.ac.uk}
$^{2}$2Dyson School of Design Engineering, Imperial College London, SW7 2DB, London, UK
}
\vspace{-0.4cm}
\maketitle
\vspace{-0.4cm}
\IEEEpeerreviewmaketitle
\vspace{-0.5cm}
\section{Introduction}
\textbf{ My research objective is to explicitly bridge the gap between high computational performance and low power dissipation of robot on-board hardware  by designing a bio-inspired tapered whisker neuromorphic 
computing ( also called  \emph{reservoir computing}) system for offroad robot environment perception and navigation, that centres the interaction between a robot’s body and its environment.}

Mobile robots performing tasks in unknown environments
need to traverse a variety of complex terrains, and they must
be able to reliably and quickly identify and characterize
these terrains to avoid getting into potentially challenging or
catastrophic circumstances\cite{otsu2016autonomous}. However, most current methods, such as  vision \cite{howard2001vision}, lidar \cite{suger2015traversability}, audio \cite{christie2016acoustics,valada2017deep}, inertial measurement units (IMU) \cite{brooks2005vibration}, tactile sensors \cite{giguere2011simple,baishya2016robust,fend2003active}, and multi-modal fusion \cite{milella2015self,reina2016lidar} ,
require a large number of training data, which is typically difficult to obtain for offroad robots.
Moreover, it involves manual labelling to build a training dataset, which is not trivial in complex, unknown environments\cite{wellhausen2019should}.
Furthermore, these methods need a significant amount of computational resources for data processing e.g, Fourier transformation and online neural network training, and their classification performance tends to deteriorate when there is no sufficient variability in the training dataset \cite{wellhausen2019should}.
As a result, quickly and cost-effectively identifying and predicting the surface characteristics of an unknown terrain in complex extreme environments  such as Mars is difficult for practical robotic applications, especially, with IMU \cite{giguere2011simple,baishya2016robust,dupont2008frequency} and vision\cite{salman2016advancing} \cite{wellhausen2019should} methods which require complicated processing of raw data, and its. 

\textbf{There is a clear and pressing need to solve the conflict between the computational power of the existing on-board hardware and the algorithmic arithmetic requirements for offroad robots,  to reach a higher level of autonomy, especially as robots become smaller and multiple tasks need to be performed locally with low latency, lower power, and high efficiency.}

To solve this problem, I drew inspiration from animals like rats and seals, just relying  on whiskers to perceive surroundings information and survive in dark and narrow environments\cite{de2022insect}. Additionally, I looked to the human cochlear which can separate different frequencies of sound. Based on these insights, 
\textbf{my work addresses this need by  exploring the  physical whisker-based reservoir computing for  quick and cost-efficient mobile robots environment perception and navigation step by step.} This research could help us understand how the compliance of the biological counterparts helps robots to dynamically interact with the environment and provides a new solution compared with current methods for robot environment perception and navigation with limited computational resources, such as Mars \cite{yu2022tapered}.
\vspace{-0.2cm}
\section{COMPLETED WORK}
Our previous research  has creatively solved a machine-learning terrain  classification and texture estimation problem by using a tapered ‘electro-mechanical’ whisker-based neuromorphic computing system for the first time in the world. I will go over my work
so far, see Fig. 1 for a map of my work.
\vspace{-0.5cm}
\subsection{Straight Whisker Sensor for Terrain Identification based on Nonlinear Dynamics}

To solve the problem we proposed before, 
we first show analytical and experimental evidence of using the vibration
dynamics of a compliant whisker based on a cylindrical spring for accurate terrain
classification during the steady-state motion of a mobile robot \cite{yu2021method}. A Hall effect sensor was used to measure whisker vibrations due to perturbations from the ground. Analytical results predict that the whisker vibrations will have one dominant frequency at the vertical perturbation frequency of the mobile robot and one with distinct frequency components. These frequency components may come from bifurcation of vibration frequency due to nonlinear interaction dynamics at steady state.  This nonlinear dynamic feature is used in a deep multi-layer perceptron neural network to classify terrains. We achieved 85.6\% prediction success rate for seven flat terrain surfaces with different textures at the speed of $0.2 $m/s.
The average prediction accuracy of the seven terrains   are: 83.31\%, 85,74\%, 85.60\%, 87.8\%, and 84.84\%, at speeds of $0.1$ m/s, $0.15 $m/s, $0.2 $m/s, $0.25$ m/s and $0.3$ m/s respectively. The experimental results  exhibit distinct dominant frequency components unique to the speed of the robot and the terrain roughness.
\vspace{-0.2cm}
\subsection{Tapered Whisker-Based Physical Reservoir Computing System for Offroad Robot Terrain
Identification }

Based on the findings in\cite{yu2021method}, we present for the first time the use of tapered whisker-based reservoir computing (TWRC) system mounted on a mobile robot for terrain classification  and roughness estimation of unknown terrain \cite{yu2022tapered}. The
results of this research provide two significant findings.
Firstly, the  tapered spring could serve as a  reservoir to map time-domain vibrations signals caused by the interaction perturbations from the ground to frequency domain features directly. In that case,  these temporal vibration signals could be processed efficiently by the proposed TWRC system which can provide morphological computation power for data processing and reduce the model training cost compared to the convolutional neural network (CNN) approaches\cite{nakajima2020physical,nakajima2015information}. We achieved a prediction success rate of 94.3\% for six terrain surface classification experiments and 88.7\% for roughness estimation of the unknown terrain surface \cite{yu2022tapered}.
The results  also revealed  that the steady response of a tapered whisker sensor could be used to identify the terrain texture even between very similar terrains, and external terrain stimuli of different roughness and hardness will produce unique reservoir features, which can be used to predict the roughness of the terrain. Therefore, an extended TWRC method including a novel detector is proposed based on the Mahalanobis distance in the Eigen space,  which has been experimentally demonstrated to be feasible and sufficiently accurate. 
\vspace{-0.1cm}
\begin{figure*}[!htb]
\centering
\includegraphics[width=0.98\textwidth]{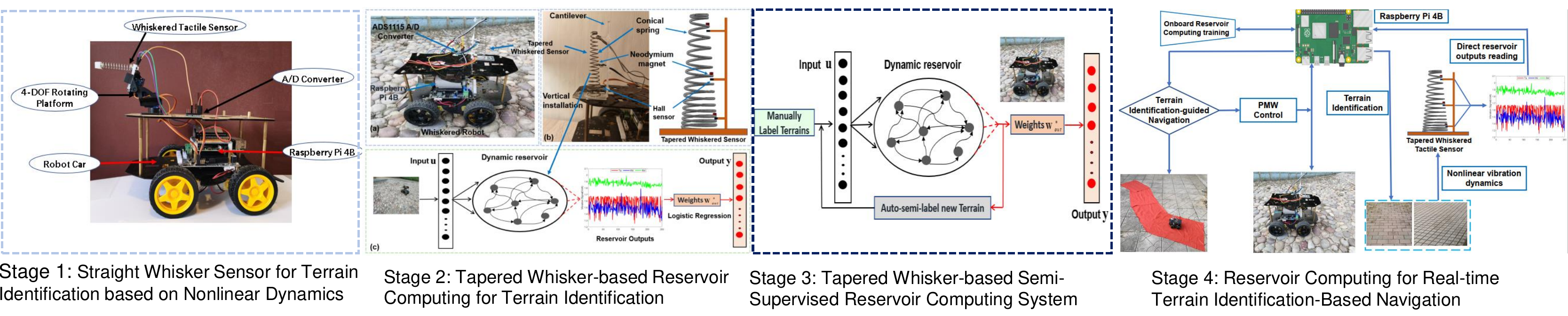}
\caption{My work has bridged the gap between high computational performance and low power dissipation of robot on-board hardware  by designing a bio-inspired tapered whisker neuromorphic 
computing system.}
\label{Animal Whisker}
\vspace{-0.5cm}
\end{figure*}
\vspace{-0.2cm}
\subsection{Semi-Supervised Whisker-based Reservoir Computing System for Offroad
Robot Terrain Texture Estimation}
\vspace{-0.2cm}
We proposed  a novel tapered whisker-based semi-supervised reservoir computing  algorithm  for the first time  for new terrain identification,  terrain texture estimation, and terrain classification \cite{yu2022semi} \cite{yu2022tapered}.  This approach was validated using a whiskered robot in an unstructured environment so that could successfully detect the new terrain classes,  and achieve  $84.12\%$ accuracy over six terrains with carpet representing the new terrain class at the speed of 0.2m/s.
The proposed scheme  could also precisely estimate the terrain properties  of new terrain and auto-label it based on prior existing experience.  For example,  the proposed method could give relatively accurate predictions when they are used as unknown terrain, e.g. $Brick = 0.75*Gravel + 0.25*Flat$, which objectively reflects the properties of the new terrain.
It is also worth noting that the system achieves its best experiments at speeds $0.4m/s$, however, the recognition accuracy decreases at speeds $0.1m/s$ and $0.5m/s$.
In all, the  experiment results  have
shown the potential of the proposed method  for the terrain identification and classification of a completely unknown world without an excessive amount of human intervention and the whiskered robot could use speed control to achieve better identification performance. Besides, this method is less affected by the vibration of the robot itself, so it can be well transferred to other robots without customization \cite{yu2021method}.
\vspace{-0.3cm}
\subsection{Tapered Whisker Neuromorphic Computing for Real-time Terrain Identification-Based Navi-
gation}
\vspace{-0.2cm}
Considering the computational superiority of neuromorphic computing,  a real-time terrain identification-based navigation method is proposed using an on-board tapered whisker-based reservoir computing (TWRC) system \cite{yu2023tapered}.
The nonlinear dynamic response of the tapered whisker was investigated in various analytical and  Finite Element Analysis (FEA) frameworks to demonstrate its reservoir computing capabilities.
Numerical simulations and experiments were cross-checked with each other to verify that whisker sensors can separate different frequency signals directly in the time domain and demonstrate the computational superiority of the proposed system and that different whisker axis locations and motion velocities provide variable dynamical response information.
Terrain surface-following experiments demonstrated that  our system could accurately identify changes in the terrain real time and adjust its trajectory  to stay on specific terrain. This interdisciplinary embodiment intelligence research could bridge the gap between high computational performance and low power dissipation  of robot on-board hardware

\vspace{-0.1cm}
\section{FUTURE WORK} 
\label{sec:conclusion}
\vspace{-0.2cm}
Moving forward, my future research interest lies in the intersection between bio-inspired robotics design, multi-sensor fusion and machine learning (especially neuromorphic computing) and their applications in the real world, such as mobile robot autonomous navigation, exploration, etc. 

Based on neuromorphic computing principles \cite{sandamirskaya2022neuromorphic,sandamirskaya2022rethinking}, I plan to  develop a neuromorphic computing sensor along with neuromorphic algorithms to iteratively discover physical hardware and embodied intelligence improvements for offroad robot terrain perception, with minimum onboard computation complexity. Neuromorphic hardware enables fast and power-efficient neural network–based artificial intelligence that is well suited to solving offroad robot terrain perception and navigation tasks. Neuromorphic algorithms can be further developed following neural computing principles and neural network architectures inspired by biological neural systems.

I believe that neuromorphic computing, featuring embodied sensing exploiting real-time information procession of the robot for different interaction terrain scenarios would introduce meaningful neuromorphic computing hardware design changes for better AI performance.


\bibliographystyle{plainnat}
\bibliography{references}

\end{document}